\pdfoutput=1
\documentclass{article}
\usepackage{spconf,amsmath,graphicx}
\usepackage{multirow}
\usepackage{mathptmx}
\usepackage[medium, compact ]{titlesec}
\usepackage[noend]{algorithmic}
\usepackage{algorithm}
\usepackage{float}

\title{
adaptive Axonal Delays 
in  feedforward spiking neural networks for accurate spoken word recognition}

\name{Pengfei Sun$^1$, Ehsan Eqlimi$^1$, Yansong Chua$^2$, Paul Devos$^1$, Dick Botteldooren$^1$\thanks{pengfei.sun@ugent.be, caiyansong@cnaeit.com} \thanks{This research is supported by the Research Foundation - Flanders under grant number G0A0220N and the Flemish Government under the "Onderzoeksprogramma Artificiele Intelligentie (AI) Vlaanderen". This research is also supported
by the National Key Research and Development Program of China (Grant No. 2021ZD0200300).}}
\address{
  $^1$WAVES Research Group, Ghent University, Belgium\\
  $^2$China Nanhu Academy of Electronics and Information Technology, China}

\begin{document}
%

\maketitle
%
\begin{abstract}
Spiking neural networks (SNN) are a promising research avenue for building accurate and efficient automatic speech recognition systems. Recent advances in audio-to-spike encoding and training algorithms enable SNN to be applied in practical tasks. Biologically-inspired SNN communicates using sparse asynchronous events. Therefore, spike-timing is critical to SNN performance. In this aspect, most works focus on training synaptic weights and few have considered delays in event transmission, namely axonal delay. In this work, we consider a learnable axonal delay capped at a maximum value, which can be adapted according to the axonal delay distribution in each network layer. We show that our proposed method achieves the best classification results reported on the SHD dataset ($92.45$\%) and NTIDIGITS dataset ($95.09$\%). Our work illustrates the potential of training axonal delays for tasks with complex temporal structures.
\end{abstract}

\section{Introduction}
Accurate and efficient speech recognition models are key to realizing automatic speech recognition (ASR) applications in low-power mobile devices and smart appliances \cite{gopalakrishnan2020hfnet}. ASR systems with advanced deep artificial neural networks (ANN)~\cite{ graves2013speech}, powerful feature extraction algorithms (e.g., MFCC, Log-Mel), and massive data have achieved great success, reaching human-level performance in some tasks. However, the improved performance comes with higher energy consumption, and the current feature extraction method is less biologically realistic~\cite{baby2021convolutional} and processes audio signals in a synchronous manner. This motivates us to explore more efficient solutions. 

Recently, biologically realistic auditory encoding frontend \cite{Wu2018, pan2020efficient, wu2018spiking} coupled with an SNN has received increasing attention. In particular, recent studies~\cite{Neftci2019, Shrestha2018, zhang2021rectified} using surrogate gradients or transfer learning \cite{wu2021tandem,wu2021progressive,yang2022training} to circumvent the problem of non-differentiable spike functions have opened the way for applying relatively deep SNN on large datasets. However, applying SNN to ASR is in its early phase, while more recent works have made progress in event-driven audio processing algorithms~\cite{wu2019robust, wu2020deep, 10.1007/978-981-10-5230-9_57, Blouw2020,yilmaz2020deep, zhang2019mpd}.

Nevertheless, most SNN studies focus on training synaptic weights~\cite{shrestha2017robust} and membrane time constants~\cite{perez2021neural, fang2021incorporating}. Several studies also propose training synaptic delays to learn the correct spike-timing~\cite{Shrestha2018,zhang2020supervised}. These methods can be regarded as encoding information in spike-times, while little research has studied the effect of different transmission delays \cite{Shrestha2018}. Sun et al.\cite{Sun2022} adopted a learnable axonal delay, which is effective in learning tasks with high temporal complexity. However, this work implements an axonal delay that is capped at a fixed maximum delay that does not adapt. 

In this paper, we propose a training scheduler that adjusts the axonal delay cap independently for each layer during training. This improves the performance in ASR tasks. Our method adopts a two-stage training strategy. In the first stage, the network is pre-trained for several epochs to obtain the initial delay distribution. In the second stage, a mechanism to determine the layer-wise delay cap is introduced that uses two tunable parameters, the sliding window size and cap fraction, to calibrate the caps and redistribute the axonal delays.

\section{Methods Description} \label{2}
\begin{figure*}[htbp]
\centering
\vspace{-0.8cm}
\hspace{-7cm}
\includegraphics[scale=0.62]{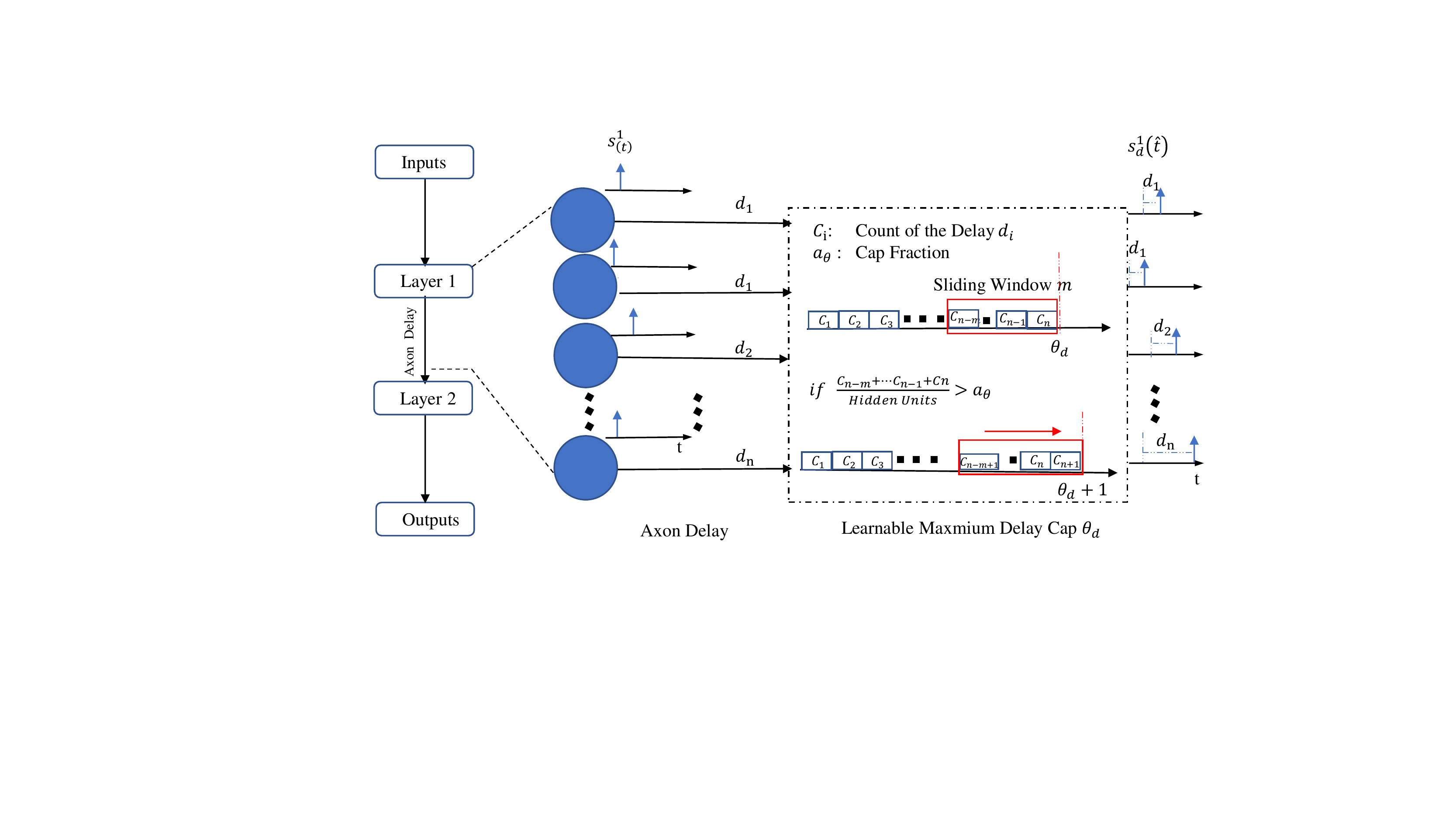}\vspace{-4cm}\hspace{-5cm}
	\caption{Illustration of how the adaptive delay caps are determined and the axonal delays adjusted. The generated spikes $s^1(t)$ will be shifted in time by $d_{i}$ and then output as spike trains $s_{d}^1(\hat t)$ in the axonal delay module. The adaptive scheduler will adjust the delay cap accordingly. The delay value may be the same across neurons, such as the top two neurons with the same delay value $d_{1}$. The layer can be a traditional convolutional layer, dense layer, or recurrent layer.}
\label{fig:input}
\end{figure*}

We first describe the spiking neuron model in sec.\ref{sec:1}, and the axonal delay module with adaptive delay cap is then introduced in sec. \ref{sec:2}.    

\subsection{Spike Response Model}\label{sec:1}
The spiking neuron model obtains its name from the special computation paradigm that communicates with spikes, which is highly energy efficient as no input integration is performed when there are no incoming events. In our work, we employ the Spike Response Model (SRM), of which membrane potential is described below
\begin{equation}
          u^{l}_{i}(t) = \sum_{j}(W^{l-1}_{ij}(\epsilon * s^{l-1}_{j})(t) + (\nu * s^{l}_{i} )(t)) 
\end{equation}
where $W_{ij}^{l-1}$ indicates the synaptic weight from neuron $j$ to $i$ at layer $l-1$ and $u^{l}_{i}$ refers to the membrane potential of neuron $i$ in layer $l$, while $s^{l-1}_{j}$ is the incoming spike pattern from the preceding neuron $j$. In this experiment, we use the response signal $a(t)=(\epsilon * s^{l-1})(t)$ to describe the response of neurons by convolving input spikes $s^{l-1}(t)$ with the response kernel $\epsilon$, where $\epsilon(t) = \frac{t}{\tau_s}\exp(1-\frac{t}{\tau_s})\Theta(t)$. Here, $\Theta(t)$ represents the Heaviside step function. Likewise, the refractory signal can be described as $(\nu * s^{l})(t)$, where $\nu(t) = -2\theta_{u}\,\frac{t}{\tau_r}\exp(1-\frac{t}{\tau_r})\Theta(t)$. Here, the parameter $\tau_s$ and $\tau_r$ are the time constant of the corresponding kernels. 
An output is generated whenever the $u^{l}_{i}$ surpasses the pre-defined threshold $\theta_{u}$. This spike-generation process can be formulated as
\begin{equation}
          s^{l}_{i}({t}) = \Theta({u^{l}_{i}(t) - \theta_{u}}) \
\end{equation}

\subsection{Axonal delay module and adaptive delay caps}\label{sec:2}
In Fig.\ref{fig:input}, the axonal delay module and an adaptive training scheduler for delay caps are shown. The axonal delay is part of spike transmission, and we formulate it such that it can be jointly learned  with an optimization algorithm. 
\begin{equation}
            s_{d}^{l}(\hat{t}) = \delta(t-{d^{l}}) * s^{l}(t)
\end{equation}
In layer $l$, $d^{l}$ represents the set of delays \{${d_{1},d_{2},..,d_{n}}$\} subject to the constraint that \{${d_{1}<d_{2},..,<d_{n}\leq \theta_d}$\}. Meanwhile, $s_{d}^{l}(\hat{t})$ denotes the spike trains output by the delay module at a shifted time $\hat{t}$. From the optimization point of view, constraining the delay value can facilitate learning. Here, we compute the fraction of neurons against the total number of neurons within a sliding window \cite{niu2019comparison} in the adaptive training scheduler so as to optimize training. 

Consider the sliding window $m$, when the fraction of delay neurons within this window exceeds the pre-defined cap fraction $\alpha_{\theta}$, the sliding window right-shifts by 1, and the delay cap $\theta_{d}$ also increases by 1. Pseudo-code of the proposed adaptive training scheduler is presented in Algorithm \ref{alg1}. 

During training (in the second while loop), the delay will be clipped as follows 
\begin{equation}
          {d} = max(0, min(d, \theta_d))
          \label{eq4}
\end{equation}
where the $\theta_d$ and delays $d$ will be adaptively adjusted according to our scheduler.
\begin{algorithm}[ht]
	\renewcommand{\algorithmicrequire}{\textbf{Input:}}
	\renewcommand{\algorithmicensure}{\textbf{Output:}}
	\caption{Pseudo-code of the adaptive training scheduler}
	\label{alg1}
	\begin{algorithmic}[0]
	\REQUIRE $\theta_d$: Delay Cap, $d$: Delay value, $\alpha_{\theta}$: Cap fraction, $m$: Sliding window size, $C_{i}$: Count of neurons with delay value $d_{i}$, $n$: Index of max delay value, $Tsteps$: Number of training steps, $Hidden Units$: Number of neurons in this layer
		\STATE {\bf Initialization}:$\left\{ {\theta_{d}} \right\},\left\{ d \right\} $
	    \STATE {\bf Pre-train model}: $x$ epochs
	    \STATE {\bf Initialization}: $n \leftarrow C.index(max(d))$, $counter \leftarrow 0$	
	    \STATE Fraction of neurons within sliding window:\\ 
	    $\alpha =\frac{(C_{n-m}+...+C_{n-1}+C_{n})}{Hidden Units}$ 
		\WHILE{{$ \alpha  > \alpha_{\theta}$ }}
		\STATE $n \leftarrow n + 1$
		\STATE $\theta_{d} \leftarrow \theta_{d} + 1$
		\STATE $counter \leftarrow 0$
		\WHILE{$counter < Tsteps$}
		\STATE Train $d$
     	\STATE Clip $d$ based on Equation~(\ref{eq4})

	\STATE $counter \leftarrow counter + 1$ 	
		\ENDWHILE
		\ENDWHILE
		\ENSURE Maximum Delay Cap $\theta_d$, Delay $d$.
	\end{algorithmic}  
\end{algorithm}

\section{Experimental Setup}\label{3}
\subsection{Datasets}
We evaluate our proposed methods on the SHD \cite{cramer2020heidelberg} and NTIDIGITS \cite{anumula2018feature} datasets. These datasets are the spike patterns processed by the artificial cochlear models\cite{cramer2020heidelberg, anumula2018feature} from the original audio signals and have high temporal complexity \cite{iyer2021neuromorphic}. The SHD dataset consists of 10,420 utterances of varying durations ($0.24s$ to $1.17s$) by  12 speakers. It contains a total of 20 digits classes from '0' to '9' in English and German language. We adopt the same data preprocessing method in \cite{yin2020effective}, and the same train/validation/test split of the dataset as the official \cite{cramer2020heidelberg}.

The NTIDIGITS is another event-based spoken recognition task. The original TIDIGITS is processd by the 64-channel CochleaAMS1b sensor and its output is recorded. In total, there are 11 spoken digits (the digits '0' to '9' in the English language and 'oh'). We follow the same train and test split as used in \cite{anumula2018feature}.
\subsection{Implementation details} 
In our experiment, we use the Adam optimizer to jointly update the synaptic weight and axonal delay with a constant learning rate of 0.1 and a minibatch size of 128. The initial caps for delay are set to 64 for SHD and 128 for NTIDIGITS, respectively. The pre-train epochs $x$ are set as $40$ and the simulation time step is 1 $ms$  for both datasets. Table 1 lists  the parameters we used in our experiments.

We train our SNN on the SLAYER-Pytorch framework \cite{Shrestha2018}. This recently introduced GPU-accelerated software is publicly available and has proven to be effective for training SNN. For both tasks, we use an SNN with two hidden fully-connected layers with 128 hidden units for SHD and 256 neurons for NTIDIGITS. The total number of model parameters is approximately 0.11M for SHD dataset and 0.08M for NTIDIGITS dataset.\\

\vspace{-0.5cm}
\begin{table}[ht]
\small
	\centering
	\caption{
Detailed parameters settings for different datasets}
	\label{tbl:param}
	\begin{tabular}{clrccc}
		\cline{1-6}
		\multicolumn{1}{c}{\bf Dataset}& \multicolumn{1}{c}{\bf $\tau_s$} & \multicolumn{1}{c}{\bf $\tau_r$} & \textbf{Initial $\theta_{d}$}  &\textbf{$\theta_{u}$}&$T_{steps}$
		\\ \hline
	\multirow{1}{*}{{SHD}}
		& 1 	& 1  & 64&10&150 \\
\multirow{1}{*}{{NTIDIGITS}}
		&  5		& 5 & 128&10&150 \\
		\hline
	\end{tabular}
	\vspace{-0.5cm}
\end{table}\textbf{}

\section{Results}\label{4}
\subsection{Ablation Study of Different Cap Fraction and Sliding Window Size}
\begin{figure}

\centering\vspace{-0.00cm}
\includegraphics[scale=0.5]{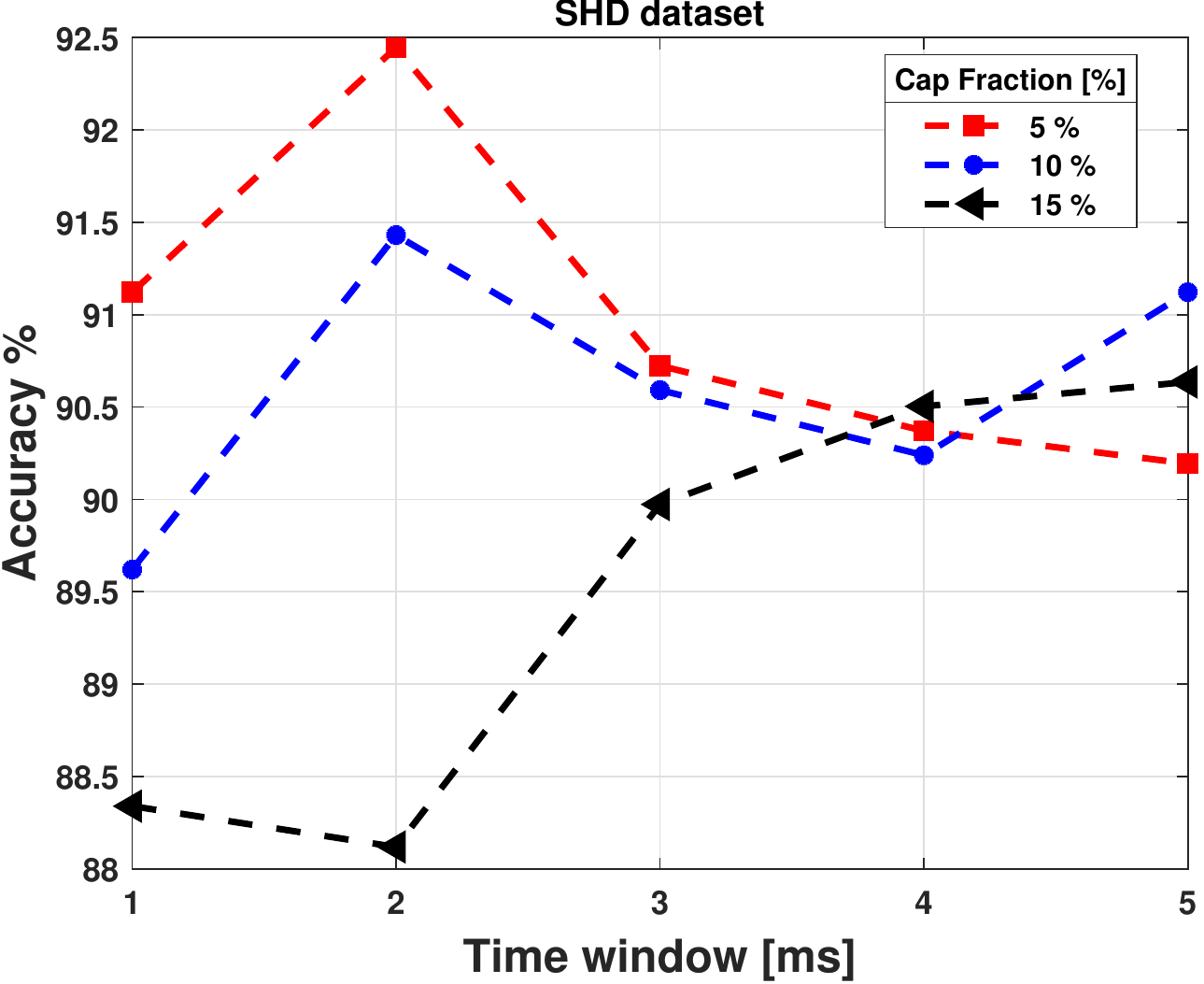}\vspace{0.2cm}
    \includegraphics[scale=0.5]{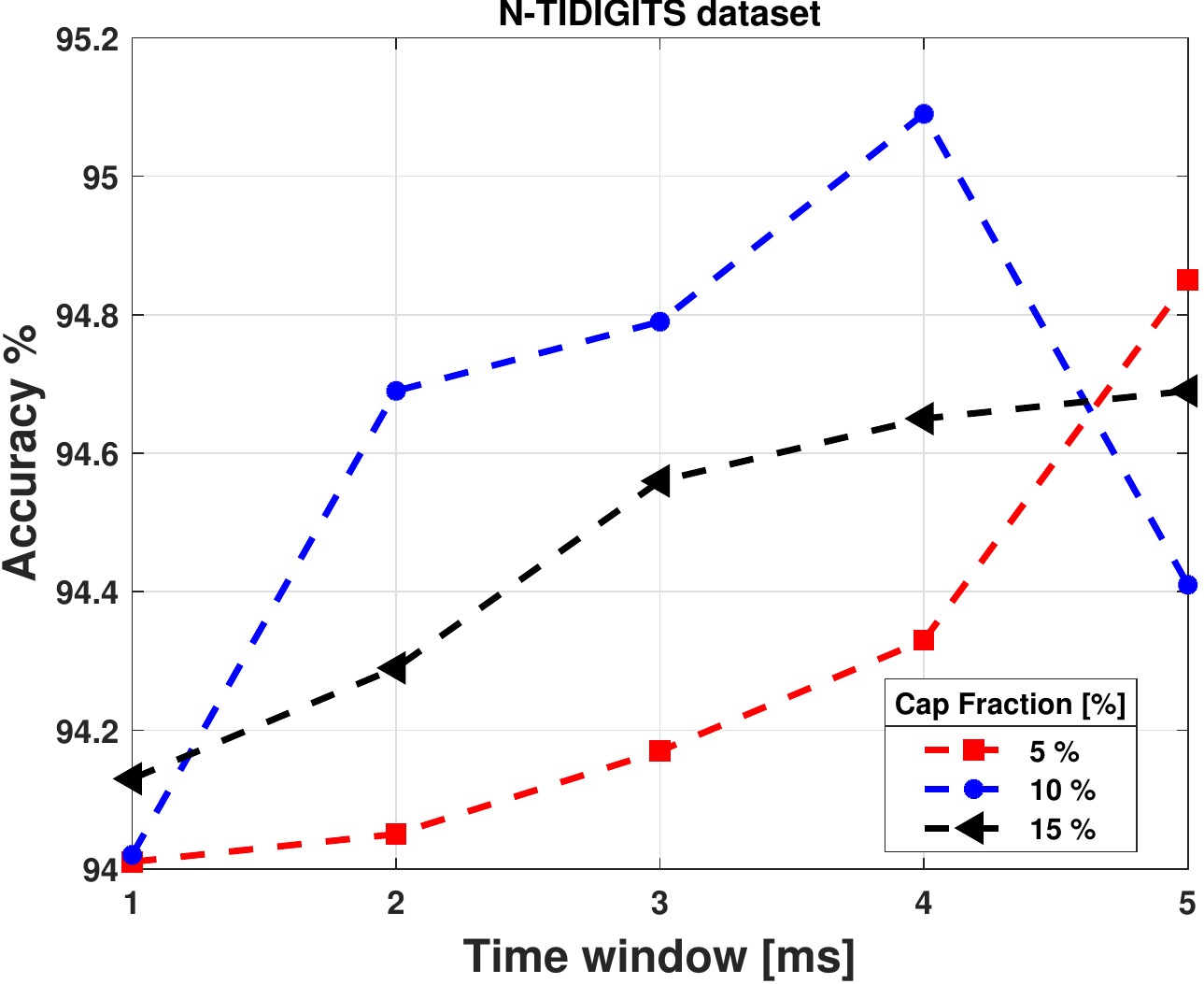}
	\caption{Ablation study of different sliding window sizes $m$ and cap fractions $\alpha_{\theta}$ based on (Top) SHD  and (Bottom) N-TIDIGITS. The x-axis indicates the sliding window size, and the y-axis refers to the accuracy. There are 3 different cap fractions  (5\%, 10\%, and 15\%) and 5 different window sizes (1,2,3,4, and 5) we use in our experiment.}\vspace{-0.0cm}
\label{fig:timewindow}
\end{figure}

The window size and cap fraction are important parameters to get a reasonable delay cap. To evaluate their influence, we design an ablation study to compare the influence, which contains 5 different window sizes and 3 different fraction parameters. The impact of these parameters on SHD and NTIDIGITS datasets is shown in Fig \ref{fig:timewindow}. For the SHD dataset, we observe that a small fraction can always get good results, and 
the best results can be obtained by controlling the number of the largest two delayed neurons within 5\% of the total number of neurons, which means the window size is set as 2 and the cap fraction is 5\%.  While for the NTIDIGITS dataset, the bigger window size and fraction are more helpful, and the accuracy keeps going up when the window size grows bigger besides the situation of window size 5 and cap fraction 10\%. 

\begin{table*}[ht]
\small
\vspace{-0.8cm}
	\centering
	\caption{Comparison with the state-of-the-art in terms of network size and accuracy.}
	\label{tbl:results}
     \setlength{\tabcolsep}{0.5mm}{
	\begin{tabular}{clrc}
		\cline{1-4}
		\multicolumn{1}{c}{Datasets}& \multicolumn{1}{c}{\bf Method} & \multicolumn{1}{c}{\bf Params} & \textbf{Accuracy}
		\\ \hline
		
		\multirow{6}{*}{{SHD}}
		&   Feed-forward SNN \cite{cramer2020heidelberg}		&  0.11 MB
		& $48.6\pm 0.9\%$\\
		&   RSNN \cite{cramer2020heidelberg}		&  
		1.79 MB& $83.2\pm 1.3\%$\\
		&   RSNN with Adaption \cite{yin2020effective} 		&  
		0.14 MB&$84.4\%$ \\
		&   Heterogeneous RSNN \cite{perez2021neural} 		&  
		0.11 MB& $82.7\pm 0.8\%$\\
		&   SNN with  Time Attention \cite{yao2021temporal} 		&  
		0.12 MB&$91.08\%$ \\	
		&  \textbf{This work  (m=2,  $\alpha_{\theta}$  = 5\%)} 		&  
		\bf0.11 MB&$\bf 92.45\%$ \\
		\hline
		\multirow{5}{*}{{NTDIDIGITS}}
		& GRU-RNN \cite{anumula2018feature}$^\dagger$
						& 
						0.11 MB& $90.90\%$ \\
		& Phased-LSTM \cite{anumula2018feature}$^\dagger$
						& 
						0.61 MB& $91.25\%$ \\
		& ST-RSBP \cite{zhang2019spike}
						& 
						0.35 MB& $93.63\pm 0.27\%$ \\
		& SNN with Axonal delay	 \cite{Sun2022}	&  
		0.08 MB& $94.45\%$  \\
		& \bf This work (m=4,  $\alpha_{\theta}$ = 10\%) 		&  
		\bf0.08 MB& $\bf95.09\%$  \\
		\hline
		\multicolumn{3}{l}{\footnotesize{$^\dagger$ Non-SNN implementation.}}
	\end{tabular}}
	\vspace{-0.5cm}
\end{table*}

\subsection{SHD}
The result of performance provided by feed-forward SNN, recurrent spiking neural network (RSNN) with adaptive firing threshold, heterogeneous RSNN, RSNN with temporal attention, and our proposed method is given in Table \ref{tbl:results}. As can be seen from Table \ref{tbl:results} , our proposed method achieves an accuracy of 92.45\%, which is the best performance reported for this task. More importantly, it can be observed that the RSNN outperforms the feed-forward SNN, which implies that the task is inherently dynamic. However, our results show that a feed-forward SNN, without recurrent connections, but with an adaptive axonal delay module, can achieve flexible short-term memory and better accuracy, using fewer parameters. 

\subsection{NTIDIGITS}
When the ANN (RNN and Phased-LSTM) are applied to the digits recognition task, they achieve an accuracy of 90.90\% and 91.25\% (see Table \ref{tbl:results}), respectively. However, these networks cannot fully exploit the advantages of sparse event-based information and have to rely on the event synthesis algorithm, effectively losing the advantage gained when processing information embedded in spike-timing. Using our adaptive axonal delay module, we can achieve the best performance of 95.09\%, and compared with Zhang et al.\cite{zhang2019spike} which  directly use the spike train level features, our model can improve  1.4\% performance while only using 23\% of parameters. 

\subsection{Effect of the Delay Value}
As shown in Table \ref{tbl:delay}, the delay cap is important to the performance of the model. For both audio classification tasks, the performance is competitive without limiting the range of delay values, demonstrating the effectiveness of the axonal delay module. However, it is still necessary to limit the delay range, and our experiments show that an appropriate delay cap will improve the classification ability of the model. Taking the SHD dataset as an example, our adaptive training scheduler can determine the optimal delay distribution, and this distribution enables the model to achieve the best performance. For other combinations of delay caps, the obtained accuracy drops, may indicate that each network structure has an optimal delay cap, but it is difficult to find the delay caps manually. Our method provides an adaptive training scheduler to search for these optimal parameters. It is worth noting that the NTIDIGITS dataset conforms to this phenomenon.

\vspace{-0.0cm}
\begin{table}[ht]
\small
	\centering
	\caption{Ablation studies for different delay cap methods and the effect of the delay cap $\theta_{d}$ in the axonal delay module. $\theta_{d_i}$ indicates the delay of $i^{th}$ layer. 'Manual' refers to using a static delay cap, while 'Adaptive' refers to using our proposed adaptive scheduler.}
	\label{tbl:delay}
	\begin{tabular}{cclrc}
		\cline{1-5}
		\multicolumn{1}{c}{\bf Dataset}&
		\multicolumn{1}{c}{\bf Method}&
		\multicolumn{1}{c}{\bf ($\theta_{d_1}$,\ $\theta_{d_2}$)} & \multicolumn{1}{c}{\bf Params} & \textbf{Accuracy}
		\\ \hline
		\multirow{5}{*}{\rotatebox{90}{SHD}}
		&Manual&  (0,\ \ \ 0)		& 
		108,820& $67.05\%$ \\
		&Manual&  (64,\ 64)		& 
		109,076& $86.84\%$ \\
		&Manual&  (128,\ 128)		& 
		109,076& $87.24\%$ \\
		&Adaptive&  (107,\ 175)		& 
		109,076& $\bf92.45\%$ \\
		&Manual&  ($+\infty$,$+\infty$) 		&  
		109,076& $84.99\%$ \\
		\hline
		\multirow{4}{*}{\rotatebox{90}{NTIDIGITS}}
		&Manual&  (0, \ \ \ 0)		&  
		85,259& $78.86\%$ \\
		&Manual& (128, \ 128)		&  
		85,771& $94.19\%$ \\
		&Adaptive& (215, \ 215)		&  
		85,771& $\bf95.09\%$ \\		
		&Manual&  ($+\infty$,$+\infty$)		&  
		85,771& $93.83\%$ \\
		\hline
	\end{tabular}
	\vspace{-0.0cm}
	\vspace{-0.5cm}
\end{table}

\section{Conclusions}\label{5}
In this paper, we integrate the learnable axonal delay module into the spiking neuron model and then introduce an adaptive training scheduler to adjust the caps of axonal delay in each network layer.  Compared to previous work that adopts a static delay cap, our proposed method significantly improves the classification capability without extra parameters. Furthermore, our adaptive scheduler can be easily integrated into the existing delay module and determine the optimal delay distribution of the network adaptively. We achieve the best performance in SHD (92.45\%) and NTIDIGITS (95.09\%) datasets with the fewest parameters. These results suggest that a neuron axonal delay with an adaptive delay cap can be used to model a lightweight flexible short-term memory module so as to achieve an accurate and efficient spoken word recognition system. We conjecture that the axonal delay mechanism introduces a form of short-term memory without increasing the number of trainable parameters. For certain data-sets in ASR, whereby 1) information is organized in short sequences, without need for long-term memory, and 2), data is limited in size and hence prone to overfitting, the axonal delay mechanism may work best, in combination with a feed-forward, small SNN. Our experiments seem to agree with the above and further confirm the great potential of using spike timing as part of the solution to an ASR problem. Furthermore, the use of spike-based losses \cite{9892379} can expedite decision-making, thereby reducing the impact of additional latency even more.  \\

\footnotesize
\bibliographystyle{IEEEbib}
\bibliography{main.bbl}


\end{document}